\documentclass[letterpaper, 10 pt, conference]{ieeeconf}  
\usepackage{FG2019}
\usepackage{graphicx,subfigure}
\usepackage{url,booktabs}
\usepackage{amsfonts}
\usepackage{tablefootnote}
\usepackage{xcolor} 

\FGfinalcopy 

\usepackage{graphics} 
\usepackage{epsfig} 
\usepackage{mathptmx} 
\usepackage{times} 
\usepackage{amsmath} 
\usepackage{amssymb}  

\def\FGPaperID{171} 

\title{\LARGE \bf
Multi-task human analysis in still images: 2D/3D pose, depth map, and multi-part segmentation
}

\author{\parbox{16cm}{\centering
    {\large Daniel S\'anchez$^1$, Marc Oliu$^1$, Meysam	Madadi$^2$, Xavier Bar\'o$^{1,2}$, Sergio Escalera$^{2,3}$}\\
    {\normalsize
    $^1$ Faculty of Computer Science, Multimedia and Telecommunication - Universitat Oberta de Catalunya, Spain \\ 
    $^2$ Computer Vision Center - Universitat Autonoma de Barcelona, Spain  \\
    $^3$ Department of Mathematics and Informatics - Universitat de Barcelona, Spain}}
}

\begin{document}


\ifFGfinal
\thispagestyle{empty}
\pagestyle{empty}
\else
\author{Anonymous FG 2019 submission\\ Paper ID \FGPaperID \\}
\pagestyle{plain}
\fi
\maketitle

\begin{abstract}
While many individual tasks in the domain of human analysis have recently received an accuracy boost from deep learning approaches, multi-task learning has mostly been ignored due to a lack of data. New synthetic datasets are being released, filling this gap with synthetic generated data.
In this work, we analyze four related human analysis tasks in still images in a multi-task scenario by leveraging such datasets. Specifically, we study the correlation of 2D/3D pose estimation, body part segmentation and full-body depth estimation. These tasks are learned via the well-known Stacked Hourglass module such that each of the task-specific streams shares information with the others. The main goal is to analyze how training together these four related tasks can benefit each individual task for a better generalization. Results on the newly released SURREAL dataset show that all four tasks benefit from the multi-task approach, but with different combinations of tasks: while combining all four tasks improves 2D pose estimation the most, 2D pose improves neither 3D pose nor full-body depth estimation. On the other hand 2D parts segmentation can benefit from 2D pose but not from 3D pose. In all cases, as expected, the maximum improvement is achieved on those human body parts that show more variability in terms of spatial distribution, appearance and shape, e.g. wrists and ankles.
\end{abstract}

\section{INTRODUCTION}

Nowadays large amounts of annotated (or weakly annotated) data are publicly available for the automatic analysis of humans \cite{c1,c4,c7,c9}. Related tasks include 2D pose estimation \cite{c2,c3,c4,c8,c9,c10}, body part segmentation \cite{c2,c4,c7,c8,c9,c10,c12}, human re-identification \cite{c8}, clothes parsing \cite{c3,c4,c12}, motion/optical flow \cite{c17,c18}, depth estimation \cite{c1, c8}, body shape model \cite{c1,c10}, body parts shape segmentation \cite{c1}, human 3D pose estimation \cite{c15,c16,c20}, or sign language recognition \cite{c44}, just to name a few.

As it is common nowadays in most computer vision problems, deep learning, and particularly Convolutional Neural Networks (CNN), is the predominant methodology used by state of the art approaches. Outstanding results have been achieved by using deep learning in tasks like 2D pose in the wild. However, other related tasks such as 3D pose, pixel-level segmentation, and human body depth estimation from RGB images still require further improvement in order to be accurately applied to real world scenarios.

Recent approaches tend to benefit from unsupervised and cross-domain scenarios \cite{c55} in order to reuse data and deal with related tasks. One standard technique in this scope is the use of multi-task approaches \cite{c7, c9, c16}. Multi-task learning has been shown to benefit human analysis tasks by leveraging the amount of data to be annotated, since each image/video does not need a full annotation of all attributes: subsets of data can be annotated for different problems. Most importantly, while solving several tasks together, information is shared among them during training, providing them with complementary information for a better generalization.

In this work we focus on multi-task learning of 2D pose, 3D pose, human body depth map, and body part segmentation from still images, which are common input cues for several human analysis tasks. Our claim is that these four tasks share semantic knowledge of the human body and, when jointly trained, can benefit each other for a better generalization. In particular, we extend the successful Hourglass network~\cite{c38} by learning each task as a separate stream and share information between tasks at different levels of the topology. Our contribution lies in the complementary analysis among the four main human body tasks on a multi-task setup. We evaluate which task combinations complement each other the best. To the best of our knowledge, this is the first time such a detailed analysis has been done in this domain.

To evaluate our work, we focus on SURREAL \cite{c1}, a synthetic dataset with realistic human bodies and annotations. Our results show that all four tasks benefit from the proposed multi-task module. We show some pairs of tasks do not help each other (e.g. 3D pose and body part segmentation), while others do so significantly (e.g. 2D pose and depth). In addition, multi-task learning provides higher performance improvements in those human body parts that show more variability in terms of spatial distribution, appearance and shape, e.g. wrists and ankles.

The rest of the paper is organized as follows. Section 2 reviews related work. Section 3 describes the multi-task network and addressed tasks. Experimental results are presented in Section 4. Finally, Section 5 concludes the paper.

\subsection{Related Work}

The use of deep-learning techniques has been a breakthrough in most computer vision applications, including human analysis scenarios. Given the need of large volumes of data to train deep learning models, there is a recent trend on learning multi-task approaches. This paradigm shares information among different tasks for a better generalization, which can leverage the amount of annotated data required for each task.

Recent works \cite{c23, c25, c26, c34} tend to extend the number of tasks to better benefit from sharing knowledge within cross-domain tasks. One extreme example can be found in \cite{c25}, where authors extend the number of tasks to eight, not just analyzing humans but objects and animals. Pyramid image decomposition is used as input to deal with semantic/boundary/object detection, normal estimation saliency/normal estimation, semantic/human part segmentation, semantic boundary detection, and region proposal generation. Other works \cite{c24, c30, c31, c32} add additional tasks such as instance segmentation, multi-human parsing, and mask segmentation. As an example, \cite{c24} tackles instance segmentation, object detection and mask segmentation in a stacked fashion. 

Different strategies exist in order to define multi-task schemes. Authors in  \cite{c55} perform a large-scale, cross-domain analysis on a new dataset of indoor scenes with no human interaction. They trained $26$ neural networks, one per category and additional combinations related to multiple domains via transfer learning. Most patterns found on this dataset exclude human kinematic constraints.
Authors in \cite{c27} build a two-stage FCN process that first detects human pose and then performs body parts parsing through a Conditional Random Field. The work of \cite{c10} uses Mask-RCNN \cite{c25} in a multi-task cascade fashion, connecting several intermediate layers for pose estimation and body parts parsing, while in \cite{c26} Mask R-CNN tackles instance/mask segmentation and object/key-point detection problems. The work of \cite{c30} makes use of adversarial networks in a nested way, i.e., GAN outputs are used as the input to other GANs to deal with pose estimation and body parts parsing. In \cite{c31,c45} recursive processing stages are used to detect and segment 2d/3d pose and body-parts. 

Another common combination of tasks is 2D/3D pose and body/clothes parsing~\cite{c16} on datasets such as Pascal~\cite{c7} or COCO~\cite{c9}. The work of \cite{c22} uses two encoders (2D pose and clothes parsing) with a module as a middle stream that acts as a parameter adapting to merge the features of both tasks and perform classification separately. In contrast, \cite{c4} proposes a two-stage multi-task procedure that first uses a residual network to extract shared features. These are used by two CNNs performing 2D pose estimation and clothes parsing, respectively.

\begin{figure}[t!]
	\centering
	\subfigure[\scriptsize  RGB]{\includegraphics[height=2.8cm]{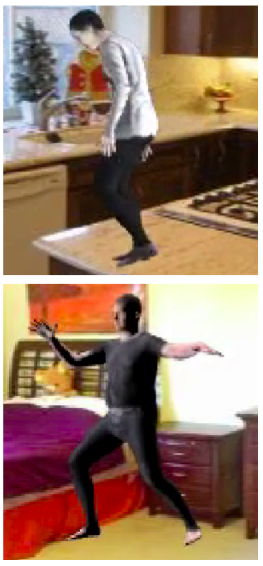}	}
	\subfigure[\scriptsize  2D pose]{\includegraphics[height=2.9cm]{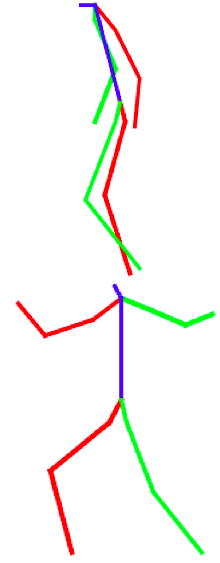}	}
	\subfigure[\scriptsize  Body Parts]{\includegraphics[height=2.8cm]{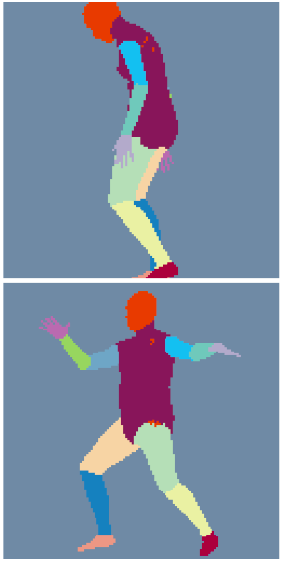}	}
	\subfigure[\scriptsize  Depth]{\includegraphics[height=2.8cm]{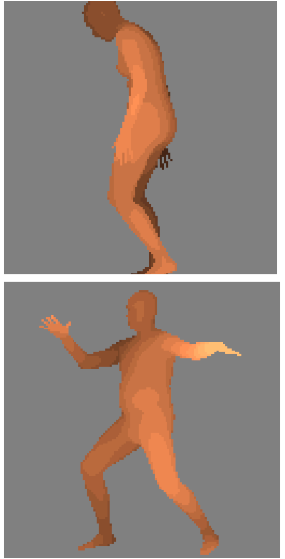}	}
	\subfigure[\scriptsize  3D pose]{\includegraphics[height=2.8cm]{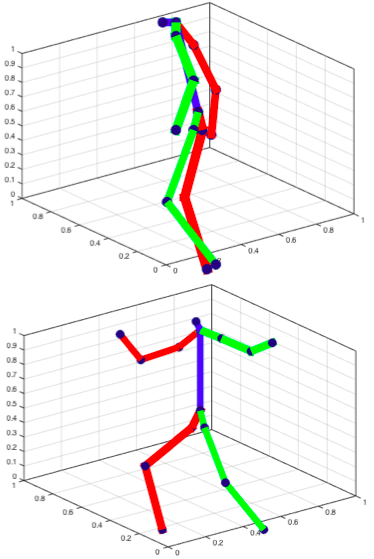}	}
    \vspace*{-2mm}
	\caption{Samples from SURREAL dataset with the chosen modalities. }
	\label{fig:dataset}
	\vspace*{-7mm}
\end{figure}

\section{Multi-task human analysis}

In this section we first address the four selected tasks and then describe the proposed multi-task architecture for this analysis. We select four common tasks in many recent works: 2D/3D pose estimation, body parts segmentation and body depth estimation. These tasks have some overlapping in the shared features/information, but each has a different definition: from depth or joints regression to pixel level classification. The goal is to design a compact model, consistent across tasks, such that overlapping features/information can be easily shared among all tasks in the model. By doing so, we can analyze which tasks are more correlated and in which parts we can achieve better improvement. The four tasks are described below.
	 
\begin{figure*}[thpb]
    \vspace*{-10mm}
	\centering
	\includegraphics[width=15.5cm,height=10cm]{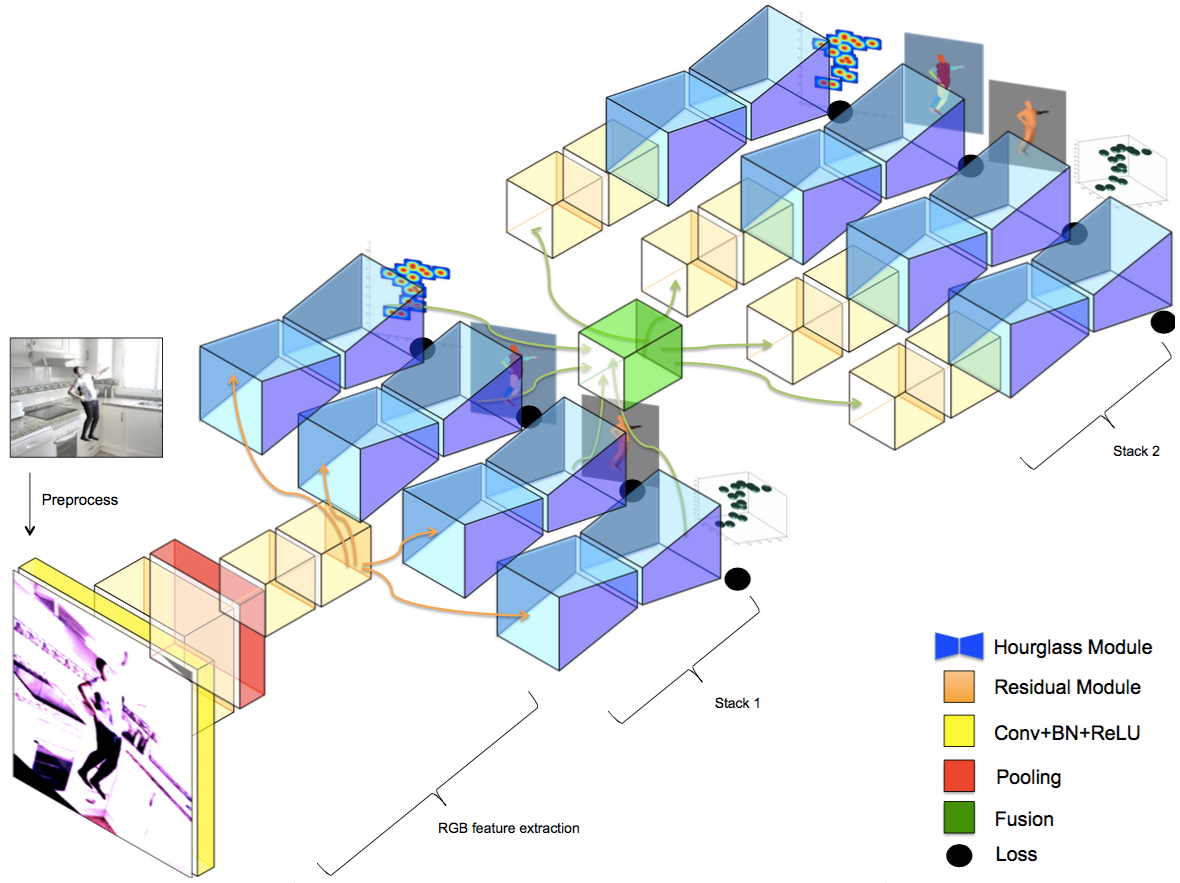}
	\vspace*{-4mm}
	\caption{Proposed multi-task architecture.}
	\label{fig:pipeline}
	\vspace*{-2mm}
\end{figure*}

\begin{itemize}
	\item \textbf{2D pose}: This task tackles the estimation of 2D human joint coordinates. Heatmaps-based methods are the state of the art for this task~\cite{c38}, consisting on estimating the location as Gaussian probability distribution around each joint. Each body joint is represented as a 2D heat map. These are stacked together, resulting in a 3D tensor where spatial relationships can be learnt~\cite{c2}. In this paper we use a tensor of size $64\times64\times16$, where $\#joints = 16$ (see Fig.\ref{fig:dataset}(b)).
	
	\item \textbf{Body parts segmentation}: The state-of-the-art on human body segmentation advocates training fully-convolutional networks that generate per pixel body part probabilities~\cite{c25,c27}. Body parts include hands, arms, legs, torso and joints like ankles and knees. We define the segmentation output as a tensor of size $64\times64\times15$ where $\#parts = 14 + Background$ (see Fig.\ref{fig:dataset}(c)). 
	
	\item \textbf{Full-body depth}: We tackle depth estimation as described in~\cite{c54}, i.e. instead of regressing each pixel depth as a continuous value we quantize depth into  $\#bins = 19$ bins resulting in a tensor of size $64\times64\times(\#bins + 1)$ (see Fig.\ref{fig:dataset}(d)). We define an extra bin for the background.

	\item \textbf{3D pose}: The standard approach for 3D pose estimation is coordinates regression~\cite{c31}. However, regressing coordinates is highly non-linear and difficult to learn by a feature-coordinates mapping~\cite{c32}. Also it is not consistent with other tasks. Following the heatmaps-based methods used in 2D pose estimation~\cite{c51}, we use the target encoding used in~\cite{c1,c23,c36}. These works encode the 3D location of the joints in the camera coordinate system~\cite{c32} into 3D heat maps. 3D Gaussians are defined by a tensor of 3 dimensions for each joint (the same number of joints as in the 2D case) taking as reference their corresponding 3D coordinates (see Fig.\ref{fig:dataset}(e)). The $x$ and $y$ axes are the standard Cartesian coordinates, being $z$-axis the depth as in the full-body depth estimation task. We output a tensor of size $64\times64\times(\#bins\times\#joints)$ by binning depth information into 19 bins for each body part.
\end{itemize}
\vspace*{-2mm}

\subsection{Multi-task architecture}
	
We define all targets at the pixel level. Therefore any fully-convolutional deep architecture can be used for individual tasks. However, in this work we consider the Stacked Hourglass network (SH)~\cite{c38}. This network has shown outstanding results for human pose estimation in still images. Each hourglass module consists of an encoder-decoder architecture with residual connections from encoder layers to corresponding decoder ones. The encoder consists of down-sampling residual modules that compress the feature space in a latent representation tensor of size $4\times4$. The decoder contains up-sampling residual modules that enlarge the tensor to $64\times64$. The residual module includes several convolutional layers plus skip connections~\cite{c37}. The skip connections from encoder to decoder allows the model to fuse low level features (e.g. edges, corners) with higher level features (e.g. semantics). The intermediate supervision at each hourglass module benefits from previous module outputs, refining and improving final network predictions. Given its high performance, its conceptual simplicity, and that allows for an easy multitask integration among stacked modules, this architecture is serving as a baseline model in several works~\cite{c47,c48,c49,c50,c51}.

In this paper we use a stream, consisting of a SH network, to learn each task. These streams are then integrated by adding intermediate connectivity and supervision, as shown in Fig.~\ref{fig:pipeline}. The resulting network is end-to-end trainable. Given an input RGB image, a set of residual modules are applied in order to generate shared features among all network streams (different tasks). Based on~\cite{c38}, several Hourglass modules can be stacked per stream. Each module has an independent supervision and provide intermediate predictions as input to the next stacks. In our case, output features from each stream are concatenated to form a tensor of size $64\times64\times(\#stream\times256)$, where 256 is the default number of Hourglass features. Next, two residual modules are applied to each stream, the first convolving the joint features to the same feature space (standard practice as shown in~\cite{c52}), and the second one compressing them to 256 features, again through convolution\footnote{Note that our contribution in this paper is not a design to compete with the state-of-the-art in each individual task, but rather a compact design to analyze cross-task contributions.}.

Regarding parameter estimation, a root-mean-square-error (RMSE) loss is used for 2D ($L_{2Dpose}$) and 3D ($L_{3Dpose}$) pose estimation, while cross-entropy (CE) across the spatial dimension of the heatmaps is used for depth estimation ($L_{Depth}$) and body part segmentation ($L_{BodyPart}$). Overall multi-task optimization is minimized by summing up the losses of all Hourglasses (\ref{eq:loss}).
\begin{equation}
L_{Total}  = L_{2Dpose} + L_{BodyPart} + L_{Depth} + L_{3Dpose}
\label{eq:loss}
\end{equation}

	
	

\section{Experiments}

Here we describe the employed dataset, metrics and analysis of all four tasks, both standalone and multi-task networks.

\subsection{Data}

In order to evaluate all multi-task combinations, we use SURREAL~\cite{c1}, a new large-scale dataset consisting of realistic synthetic data. The dataset is created by using recorded motion capture (MoCap) data to mimic realistic body movements in short video clips. The human body is rendered based on a body shape model. Then a cloth texture is added to the model including different lighting conditions. Finally, the model is projected to the image plane with a static background to have a realistic RGB image. The background is selected from indoor image datasets.
Given this synthesis pipeline, different targets can be generated along with the RGB image: body depth maps, 2D/3D coordinates, body part segmentations, optical flow and surface normals.
The dataset contains nearly 6.5M frames. It consists of 145 subjects (115 train/30 test), 2,607 (1964 train/703 test) video sequences and 67,582 clips (55,001 train/12,528 test). Some samples are shown in Fig.~\ref{fig:dataset} for the different data modalities.

\subsection{Implementation details}

We train different multi-task SH architectures considering different combinations of modalities to better analyze their complementarity. We train all models for 30 epochs using 2 Stacks of Hourglass, with a batch size of 5 and the RMSprop optimizer with learning rate $1e-3$. We first crop the image regions containing the centered human bodies using the provided bounding boxes of the dataset and resize them to $256\times256$ for training. Then, we apply standard data augmentation techniques such as scaling, jittering and rotation~\cite{c38}. Moreover, the train/test splits are done such that 20\% from the total is keep apart as in~\cite{c1}.

In order to evaluate each modality, we make use of standard metrics: Intersection over Union (IOU) for body part segmentation, Percentage of Correct Keypoints thresholded at 50\% of the head length (PCKh)~\cite{c2} for 2D pose estimation, root-mean-square-error (RMSE) for full body depth estimation and mean joint distance MJD in millimeters (mm) for 3D pose estimation. We also use success rate trend to analyze evolution of the error/accuracy within different thresholds. This is given by the percentage of frames with an error smaller than the given thresholds.
	
	\subsection{Analysis of single-task models}
	Here we evaluate the models trained on specific tasks, which will serve as baselines to multi-task comparison.
	
	\subsubsection{Body part segmentation} The first column in Table~\ref{iou} shows the single-task segmentation results, with an average IOU $67.48\%$. When looking at different body parts, the model shows high variability in accuracy: high performance for upper-body parts such as the head, torso and legs, and lower performance for the feet, upper arms and hands. This low accuracy in some parts (feet, hands) is due to these spanning just a few pixels, and regions of difficult interpretation, such as complex self-occlusions.

	\subsubsection{2D pose estimation} Regarding 2D pose estimation, the single-task 2D pose model already obtained an outstanding accuracy of $96.50\%$ PCKh, as shown in Table~\ref{2dposeTable}. This may hint to the dataset being relatively simple for this kind of task, given the current state-of-the-art approaches. More specifically, we see lower accuracy on the wrists and elbows. These need a finer location since they have large scale variations. They may also be confused with the background on cluttered environments, depending on the clothing.
	
	\subsubsection{Full-body depth estimation} As shown in Table~\ref{rmsepartsTable} the single-task depth model is capable of estimating the full-body depth (Mean Full Body row) with a $4.39\%$ RMSE, a very low error. We can measure the depth prediction error on each body part by masking the predictions with the body part segmentation masks. Results obtained using only depth (Table~\ref{rmsepartsTable}, first column) show a higher error on hands and feet, and lower error on torso, upper legs and upper arms due to their highly unconstrained kinematics in humans.
	
	\subsubsection{3D pose} In the case of 3D pose (Table~\ref{3dposetable}, first column), we obtain an average error of $60.13$mm, with the error being higher for the ankles and wrists. This is due to these covering a small spatial region, as well as corresponding to parts with many degrees of freedom. In summary, ankle, wrist and elbow are the most difficult joints to learn. Again, we see those body parts and joints are difficult to predict for all tasks.

	\begin{table*}
		\centering
		\caption{Results on SURREAL dataset measuring body parts segmentation under IOU metric}
		\vspace*{-2mm}
		\label{iou}
		\resizebox{\textwidth}{!}{%
			\begin{tabular}{|c|c|c|c|c|c|c|c|c|}
				\hline
				IOU & seg. & seg. + depth & 3D pose + seg. & 2D pose + seg. & 2D pose + seg. + depth & 2D/3D pose + seg. & 3D pose + seg. + depth & 2D/3D pose + seg. + depth \\ \hline
				Background & 98.0329 & 98.0726 & 98.0012 & 98.0631 & \textbf{98.0781} & 98.0641 & 98.0732 & 97.7579 \\ \hline
				Head & 74.3689 & 74.4037 & 73.7297 & 74.2553 & 74.1704 & 74.3771 & 74.2328 & \textbf{74.7454} \\ \hline
				Torso & 84.6390 & 84.8324 & 84.3057 & \textbf{84.9853} & 84.8013 & 84.9098 & 84.9153 & 80.6780 \\ \hline
				Upper R.Arm & 65.8220 & 66.6616 & 65.8635 & 67.0540 & 66.1376 & 66.7216 & 66.4946 & \textbf{67.7473} \\ \hline
				Lower R.Arm & 62.0338 & 62.5079 & 61.2258 & 62.6833 & 62.5857 & 62.1622 & 62.9103 & \textbf{63.0192} \\ \hline
				R. Hand & 49.3243 & 48.4630 & 48.2553 & 49.4606 & 50.8114 & 48.5266 & 48.7750 & \textbf{50.5932} \\ \hline
				Upper L.Arm & 65.4599 & 66.2077 & 65.8938 & 66.2191 & 65.3423 & 65.5865 & 65.8665 & \textbf{66.7359} \\ \hline
				Lower L.Arm & 60.5462 & 61.4842 & 61.1205 & 61.1449 & 61.1934 & 60.5194 & 61.5981 & \textbf{61.8868} \\ \hline
				L.Hand & 48.9188 & 48.9596 & 48.3028 & 46.8697 & \textbf{49.2583} & 46.1885 & 46.8591 & 48.6889 \\ \hline
				Upper R.Leg & 75.2125 & 76.1054 & 75.1161 & 76.0184 & \textbf{76.1120} & 75.9127 & 75.8433 & 76.0172 \\ \hline
				Lower R.Leg & 71.4514 & 72.2720 & 71.0750 & 71.9844 & \textbf{72.3200} & 71.7808 & 71.8441 & 71.9378 \\ \hline
				R.Feet & 55.2237 & 55.5759 & 54.5336 & 55.4427 & \textbf{56.7137} & 54.4733 & 56.3635 & 55.8747 \\ \hline
				Upper L.Leg & 75.2612 & 75.7805 & 75.4932 & \textbf{76.2944} & 76.2151 & 75.7273 & 76.1662 & 75.7628 \\ \hline
				Lower L.Leg & 71.4049 & 72.2354 & 71.0951 & 72.2172 & \textbf{72.3119} & 71.5317 & 72.1089 & 71.5965 \\ \hline
				L.Feet & 54.6201 & 55.1762 & 53.5035 & 53.9172 & \textbf{56.4636} & 53.6977 & 55.0263 & 54.9013 \\ \hline
				\textbf{Mean} & 67.4880 & 67.9159 & 67.1677 & 67.7740 & \textbf{68.1677} & 67.3453 & 67.8051 & 67.8629 \\ \hline
			\end{tabular}%
		}
		\vspace*{-3mm}
	\end{table*}
		\begin{table*}
		\centering
		\caption{Results on SURREAL dataset measuring 3D pose under MJD (mm) metric}
		\vspace*{-4mm}
		\label{3dposetable}
		\resizebox{\textwidth}{!}{%
			\begin{tabular}{|c|c|c|c|c|c|c|c|c|}
				\hline
				MJD (mm) & 3D pose & 2D/3D pose & 3D pose + seg. & 3D pose + depth & 2D/3D pose + seg. & 2D/3D pose + depth & 3D pose + seg. + depth & 2D/3D pose + seg. + depth \\ \hline
				R.Ankle & 86.1138 & 89.1075 & 81.6803 & 83.0312 & 87.9071 & 83.4697 & \textbf{79.6775} & 90.4500 \\ \hline
				R.Knee & 59.9885 & 58.7382 & \textbf{54.4890} & 55.2095 & 56.9172 & 55.7307 & 55.1506 & 57.0098 \\ \hline
				R.Hip & 25.6693 & 26.4384 & 25.7580 & 26.0962 & 26.5351 & 25.8593 & 25.5101 & \textbf{25.4791} \\ \hline
				L.Hip & 25.4341 & 25.6198 & 25.7240 & 25.5058 & 26.2216 & 25.4403 & 25.5606 & \textbf{25.0999} \\ \hline
				L.Knee & 56.9181 & 59.8854 & 56.5708 & 56.4425 & 58.4527 & 55.4873 & \textbf{55.3666} & 57.2093 \\ \hline
				L.Ankle & 87.7192 & 89.7298 & 82.2631 & 84.6840 & 86.5020 & 83.1461 & \textbf{81.0259} & 87.6353 \\ \hline
				Thorax & 31.2580 & 31.3804 & 31.4161 & 30.9884 & 31.1042 & 31.3439 & 30.5244 & \textbf{30.0228} \\ \hline
				Upper Neck & 44.5032 & 42.5916 & 42.7647 & 42.3535 & 42.1803 & 42.7474 & \textbf{41.2902} & 42.2552 \\ \hline
				Head Top & 49.6059 & 47.1529 & 46.9462 & \textbf{46.7176} & 49.1450 & 47.1224 & 46.8806 & 47.2783 \\ \hline
				R.Wrist & 103.3092 & 103.4721 & 101.2466 & 107.9753 & 107.3964 & 105.2247 & \textbf{100.6127} & 102.6424 \\ \hline
				R.Elbow & 70.3126 & 71.3751 & 70.3185 & 74.4315 & 72.5787 & 70.6057 & \textbf{68.8732} & 70.5880 \\ \hline
				R.Shoulder & 46.1421 & 45.4316 & 46.1537 & 45.6363 & 45.7330 & 44.8304 & \textbf{43.3576} & 44.1882 \\ \hline
				L.Shoulder & 47.4316 & 47.8410 & 45.2717 & 45.5204 & 46.2654 & 44.9013 & \textbf{43.9592} & 45.5271 \\ \hline
				L.Elbow & 67.3347 & 68.6716 & 67.8447 & 68.5134 & 68.7963 & 67.6302 & \textbf{63.9457} & 65.2997 \\ \hline
				L.Wrist & 100.3381 & 99.5600 & 96.5120 & 102.8758 & 103.0282 & 100.1062 & \textbf{93.1507} & 93.9053 \\ \hline
				\textbf{Mean} & 60.1386 & 60.4664 & 58.3306 & 59.7321 & 60.5842 & 58.9097 & \textbf{56.9924} & 58.9727 \\ \hline
			\end{tabular}%
		}
	\vspace*{-3mm}	
	\end{table*}

    \vspace*{-2mm}
	\subsection{Analysis of multi-task models}
	
	The various considered tasks are highly related to each other, and are based on similar visual cues. Thus, features extracted to solve a task may help solving the others by providing a richer description of the body appearance. In this section we evaluate how multi-task models help improve the accuracy of each individual task.

	\subsubsection{Body Part Segmentation}
	As shown in Table \ref{iou}, the tasks contributing the most to body part segmentation are 2D pose and depth estimation. Training a model to jointly solve these three tasks supposes a $1\%$ improvement to the segmentation accuracy in terms of IOU (from 67.48\% to 68.16\%). Possible reasons are: 2D pose estimation may help to disambiguate pixel labels in the segmentation task by providing rough estimates of the body part locations; and depth estimation can help mitigating effects such as foreshortening, crowding and occlusion. Separately, both 2D pose and depth estimation improve the segmentation results relative to both IOU and pixel error.
	
	Table~\ref{iou} also shows that 3D body pose estimation is a poor complement for the segmentation task in terms of IOU. This may be due to the complexity of estimating the landmarks depth, with the model dedicating most of its capacity to this subtask. Moreover, the model encodes a relatively poor representation of the landmark locations in the image plane. This hypothesis is reinforced by the results of performing 2D+3D pose estimation along with body part segmentation. While 2D pose estimation does help the segmentation task, further adding 3D pose estimation results in worse accuracy than performing body part segmentation alone. The same effect happens with depth estimation and 3D pose. While depth estimation improves the overall segmentation accuracy, further performing 3D pose recovery results in worse accuracy.

	Looking at body parts results, one can see that performing 2D pose recovery along with body part segmentation improves IOU for torso, arms and legs. This is better reflected in the results for the model exploiting all considered subtasks. While adding 3D body pose recovery to the pipeline worsens the overall results of the best model, it does improve the segmentation accuracy of those parts it has been shown to improve on its own such as arms and hands.

	Overall, we can say that the cues of 2D pose and depth estimation help to improve the segmentation accuracy. At the same time, 3D pose estimation worsens the overall results but helps improve the results for some specific body parts. The best overall model is found by performing 2D pose and depth estimation along with segmentation.
	
	\subsubsection{2D pose estimation}
    The results in Table~\ref{2dposeTable} show the performance of the different multi-task models on 2D human pose estimation. We can see all task combinations improve on the single-task model, with the best results achieved by considering all tasks. Specifically, using all tasks results in a $0.51\%$ improvement on the PCKh, going from $96.50\%$ with the single-task model, to $97.01\%$ when using all tasks.

\begin{figure*}[t!]
	\centering
	\subfigure[\scriptsize  3D pose]{\includegraphics[width=0.21\textwidth]{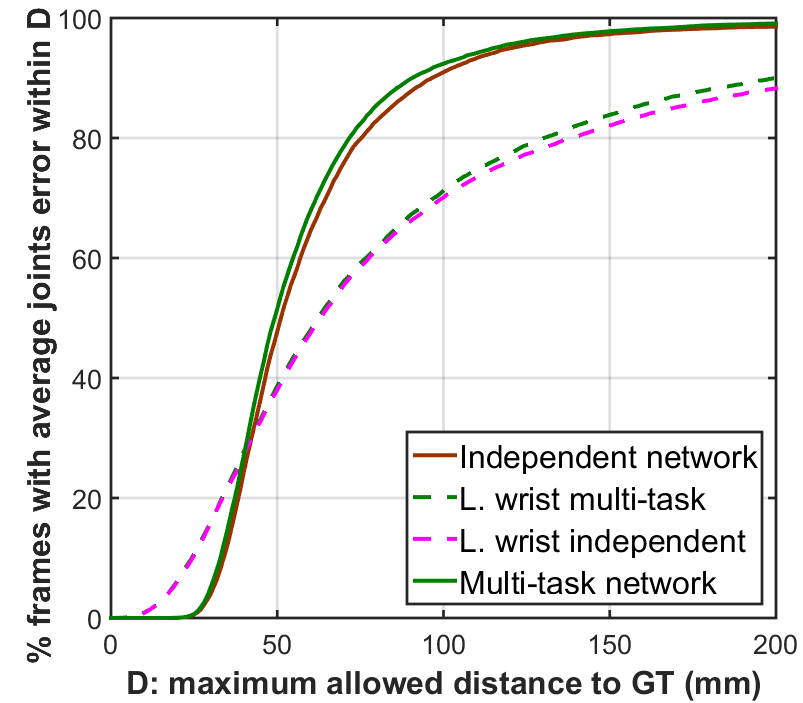}	}
	\subfigure[\scriptsize  2D pose]{\includegraphics[width=0.21\textwidth]{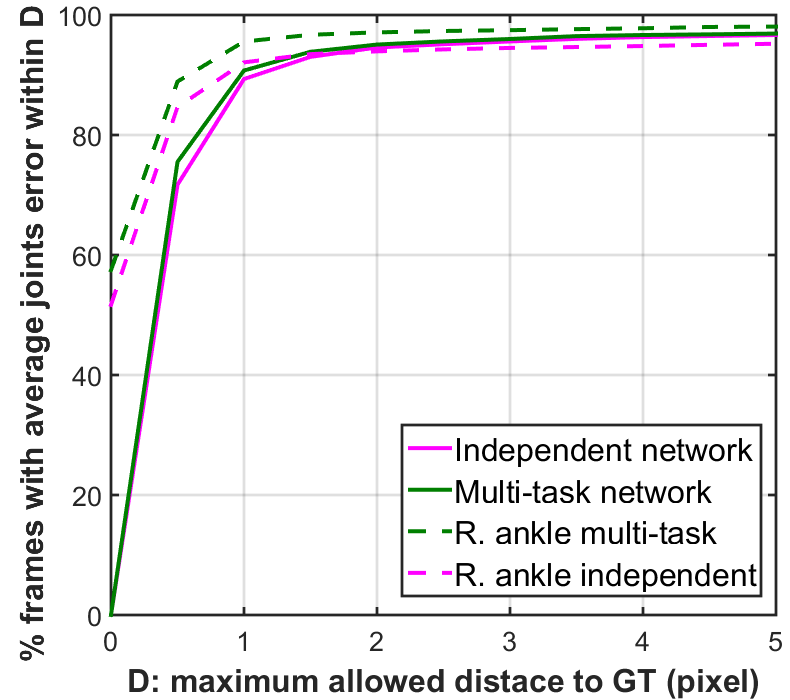}	}
	\subfigure[\scriptsize  Full-body Depth map]{\includegraphics[width=0.21\textwidth]{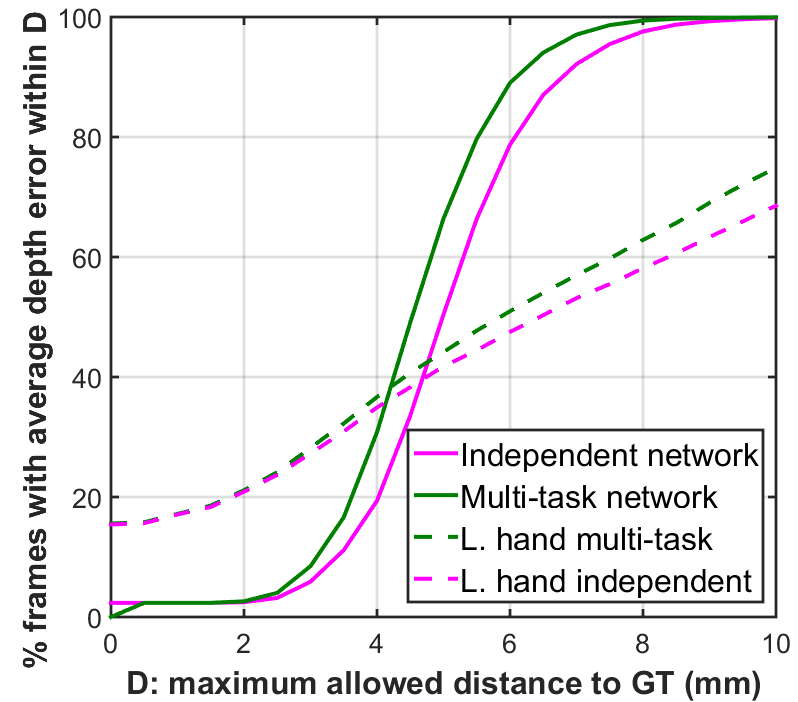}	}
	\subfigure[\scriptsize  Body Parts segmentation]{\includegraphics[width=0.21\textwidth]{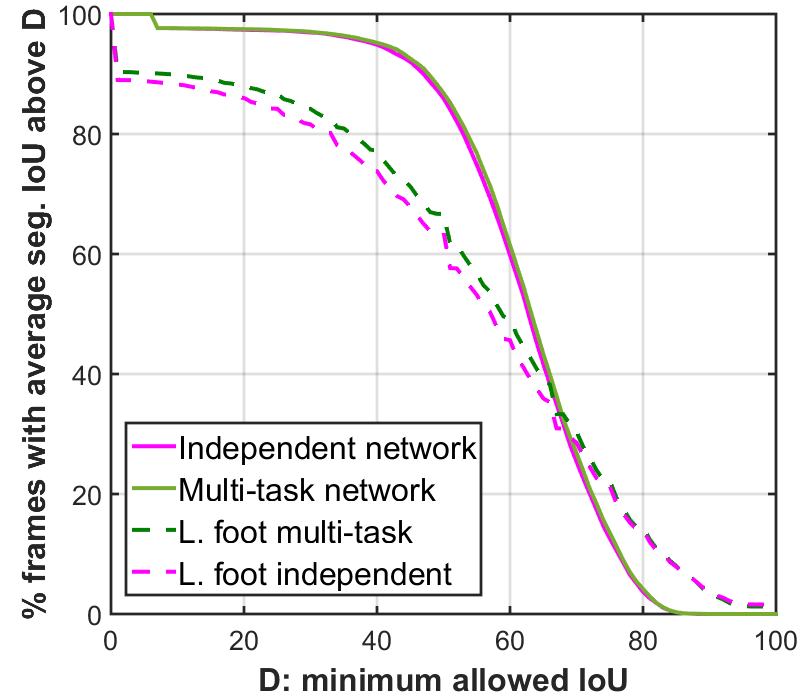}	}
	\vspace*{-3mm}
	\caption{Success rate error for the different tasks. For each task: isolated task vs best multi-task approach; and for joint/part with highest multi-task improvement, its isolated task vs multi-task score.}
	\label{fig:success_rate}
\end{figure*}

    \begin{table*}
    \centering
    \caption{Results on SURREAL dataset measuring 2D pose under PCKh metric}
    \vspace*{-3mm}
    \label{2dposeTable}
    \resizebox{\textwidth}{!}{
        \begin{tabular}{|c|c|c|c|c|c|c|c|c|}
        \hline
        PCKh & 2D pose & 2D pose + depth & 2D/3D pose & 2D pose + seg. & 2D pose + seg. + depth & 2D/3Dpose + seg. & 2D/3D pose + depth & 2D/3D pose + seg. + depth \\ \hline
        R.Ankle & 95.8064 & 95.8146 & 95.8064 & 96.3378 & \textbf{96.8119} & 96.6566 & 96.1007 & 96.6402 \\ \hline
        R.Knee & 97.0326 & 96.959 & 96.91 & 97.2942 & \textbf{97.4822} & 97.1471 & 97.0817 & 97.4495 \\ \hline
        R.Hip & 99.0109 & 99.1090 & 99.0272 & 99.1417 & 99.1090 & 99.0599 & \textbf{99.1662} & 99.0844 \\ \hline
        L.Hip & 99.1253 & 99.2725 & 99.1008 & 99.2806 & \textbf{99.3052} & 99.2234 & 99.2970 & 99.2970 \\ \hline
        L.Knee & 97.376 & 97.1552 & 97.2615 & 97.5721 & \textbf{97.8256} & 97.5149 & 97.5313 & 97.6457 \\ \hline
        L.Ankle & 96.3214 & 96.0762 & 96.3132 & 96.8364 & \textbf{97.0653} & 96.8691 & 96.4686 & 96.9427 \\ \hline
        Pelvis & 99.4687 & 99.5340 & 99.4932 & 99.5831 & \textbf{99.6158} & 99.4687 & 99.5586 & \textbf{99.6158} \\ \hline
        Thorax & 99.3787 & 99.5095 & 99.4196 & 99.5014 & 99.5177 & 99.3951 & 99.5177 & \textbf{99.5831} \\ \hline
        Upper Neck & 99.0763 & 99.1580 & 99.0763 & 99.1989 & 99.1171 & 99.0844 & 99.0763 & \textbf{99.2234} \\ \hline
        Head Top & 98.7738 & 98.8556 & 98.8065 & 98.8229 & 98.8147 & 98.7329 & 98.8474 & \textbf{98.9210} \\ \hline
        R.Wrist & 88.9970 & 89.0378 & 89.3567 & 89.9779 & 90.2068 & 89.8635 & 89.2586 & \textbf{90.6237} \\ \hline
        R.Elbow & 94.3023 & 93.5339 & 94.0898 & 94.4004 & \textbf{94.6211} & 94.4167 & 93.9671 & 94.3268 \\ \hline
        R.Shoulder & 98.3569 & 98.1362 & 98.3324 & 98.3978 & 98.5776 & 98.3733 & 98.2833 & \textbf{98.7411} \\ \hline
        L.Shoulder & 98.0136 & 98.0544 & 97.9890 & 98.1852 & 98.2343 & 98.0953 & 97.8256 & \textbf{98.4060} \\ \hline
        L.Elbow & 93.9099 & 94.2287 & 94.1061 & 94.6211 & 94.6129 & 94.6129 & 94.2042 & \textbf{94.7029} \\ \hline
        L.Wrist & 89.1850 & 89.6264 & 89.3812 & 90.2232 & 90.5420 & 90.1659 & 90.2068 & \textbf{91.0978} \\ \hline
        \textbf{Mean} & 96.5085 & 96.5039 & 96.5294 & 96.8360 & 96.9662 & 96.7925 & 96.6495 & \textbf{97.0188} \\ \hline
        \end{tabular}
        }
        \vspace*{-3mm}
    \end{table*}
    
    \begin{table*}
    \centering
    \caption{Results on SURREAL dataset measuring depth body parts estimation under RMSE metric}
    \vspace*{-3mm}
    \label{rmsepartsTable}
    \resizebox{\textwidth}{!}{%
        \begin{tabular}{|c|c|c|c|c|c|c|c|c|}
        \hline
        RMSE & depth & 2D pose + depth & seg. + depth & 3D pose + depth & 2D pose + seg. + depth & 2D/3D pose + depth & 3D pose + seg. + depth & 2D/3D pose + seg. + depth \\ \hline
        Background & 0.5151 & \textbf{0.4955} & 0.6372 & 0.5425 & 0.5727 & 0.6887 & 0.6830 & 0.5590 \\ \hline
        Head & 4.7828 & 4.8319 & 4.5270 & 4.4978 & 4.5397 & 4.3778 & \textbf{4.1174} & 4.3523 \\ \hline
        Torso & 2.7179 & 2.7216 & 2.4842 & 2.5810 & 2.538 & 2.5024 & \textbf{2.3779} & 2.5559 \\ \hline
        Upper R.Arm & 3.8742 & 3.9463 & \textbf{3.4306} & 3.8128 & 3.5647 & 3.5505 & 3.4756 & 3.4641 \\ \hline
        Lower R.Arm & 5.4385 & 5.4198 & 5.1384 & 5.3129 & 4.9385 & 5.1128 & \textbf{4.8428} & 5.0613 \\ \hline
        R. Hand & 7.0447 & 7.0778 & 7.0683 & 6.9167 & 6.6483 & 6.8738 & \textbf{6.6380} & 6.9056 \\ \hline
        Upper L.Arm & 3.7487 & 3.9582 & 3.4299 & 3.7295 & 3.5149 & 3.4873 & \textbf{3.3240} & 3.3965 \\ \hline
        Lower L.Arm & 5.4778 & 5.6605 & 5.2851 & 5.4003 & 5.0954 & 5.1793 & \textbf{4.8899} & 5.1538 \\ \hline
        L.Hand & 7.1597 & 7.2365 & 7.1001 & 6.9587 & 6.7485 & 6.9643 & \textbf{6.6202} & 6.9522 \\ \hline
        Upper R.Leg & 3.3767 & 3.4739 & 3.2649 & 3.4919 & 3.2430 & 3.3522 & \textbf{3.1933} & 3.3732 \\ \hline
        Lower R.Leg & 5.2455 & 5.3893 & 5.4107 & 5.3243 & 5.0982 & 5.1117 & \textbf{4.8619} & 5.1820 \\ \hline
        R.Feet & 7.8622 & 7.9182 & 8.0064 & 7.9454 & 7.4462 & 7.7262 & \textbf{7.3937} & 7.7420 \\ \hline
        Upper L.Leg & 3.3694 & 3.5158 & 3.2660 & 3.4426 & 3.2235 & 3.3606 & \textbf{3.2014} & 3.3314 \\ \hline
        Lower L.Leg & 5.1918 & 5.4304 & 5.4314 & 5.3661 & 5.1769 & 5.1402 & \textbf{4.9026} & 5.1566 \\ \hline
        L.Feet & 7.8774 & 8.0233 & 8.0535 & 7.9773 & 7.6125 & 7.8496 & \textbf{7.5477} & 7.8853 \\ \hline
        \textbf{Mean Body Parts} & 4.9122 & 5.0066 & 4.8356 & 4.8867 & 4.6641 & 4.7518 & \textbf{4.5378} & 4.7381 \\ \hline
        \textbf{Mean Full Body} & 4.3900 & 4.2300 & 4.3100 & 4.3500 & 4.1900 & 4.2500 & \textbf{4.0400} & 4.2400 \\ \hline
        \end{tabular}%
        }
    \end{table*}

		The single task contributing the most to 2D pose recovery is segmentation, resulting in $0.3\%$ increase. Said task may provide cues for the exact outline and localization of body parts, which can be easily leveraged for 2D body pose recovery. This is not the case of depth estimation, where body parts are not segmented. Still, depth estimation slightly improves the results, likely due to it providing an outline of the overall body, along with depth cues of said outline, helping to disambiguate the location of the parts. 3D pose estimation, on the other hand, provides little complementary information about the landmarks location relative to the camera plane, if any at all. If we look at individual joints, combining 2D pose, segmentation and depth improves on ankles and knees. Combining 2D/3D pose, segmentation and depth improves on the upper body and upper legs at the expense of losing precision on the other joints. This trade-off may be due to the ability of 3D pose estimation to disambiguate those joint locations suffering from cluttering and occlusions.
    	
    Summarizing, we see that performing all 4 tasks obtains the best results. By analyzing the other task combinations, we see that segmentation helps the most, followed by depth estimation. Finally, 3D body pose estimation only helps marginally.
    
	\subsubsection{Full-body depth estimation}
    	Here we evaluate the error on depth estimation for a collection of multi-task networks. Specifically, Table~\ref{rmsepartsTable} shows that complementing depth estimation with 3D pose estimation and body part segmentation results in the best results: while the single-task model obtains a mean $4.39$ RMSE calculated directly from the full-body depth prediction, the multi-task model goes down to a RMSE of $4.04$, an $8\%$ error reduction. Mean Body parts is the average of computing RMSE at each body part using its segmentation masks. Looking at tasks individually, segmentation contributes the most, with 3D pose estimation following closely. Segmentation may help depth estimation by providing richer semantic information on the body parts being segmented, allowing for a better model of the possible depth variability. On the other hand, 2D pose estimation does not contribute to solving the task, resulting in a higher error. This is due to this task not making use of depth information, resulting in a bigger combined feature space with no additional depth cues in the encoding. We see this in higher order combinations: combining the successful tasks (segmentation and 3D pose estimation in addition to depth estimation) results in the best results. Further adding 2D pose estimation to the pipeline increases the overall error.
	
	If we look at the results by body part (Table~\ref{rmsepartsTable}), the best model, combining all tasks except for 2D pose estimation, obtains the lowest error in all cases. Compared to the baseline, some improvements to remark are head, lower arms and hands. This is due to the contribution of segmentation to better localize the parts layout and the 3D pose information to refine ambiguities at the depth level. Some difficult parts include the feet, lower legs and hands.

	\begin{figure*}[thpb]
	\centering
	\subfigure[\scriptsize Segmentation IOU:\newline seg.  vs 2D+seg.+depth]{\includegraphics[height=2.8cm]{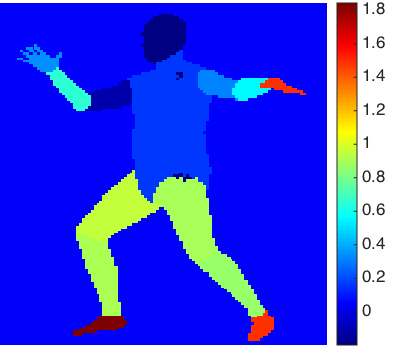}}
	\subfigure[\scriptsize  Segmentation Pixel Accuracy: seg. vs 2D+seg+depth]{\includegraphics[height=2.8cm]{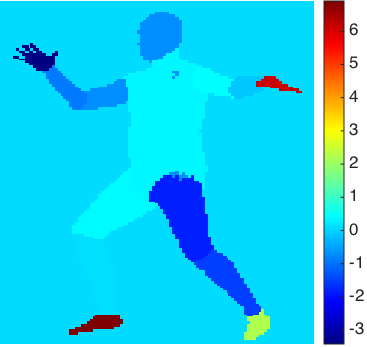}	}
	\subfigure[\scriptsize  Full-body depth:\newline depth vs 3D+seg.+depth]{\includegraphics[height=2.8cm]{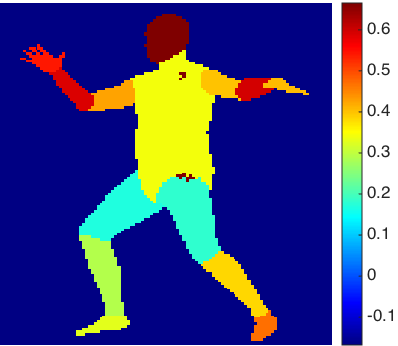}}	
	\subfigure[\scriptsize  2D pose:\newline 2D vs 2D/3D+seg.+depth]{\includegraphics[height=2.8cm]{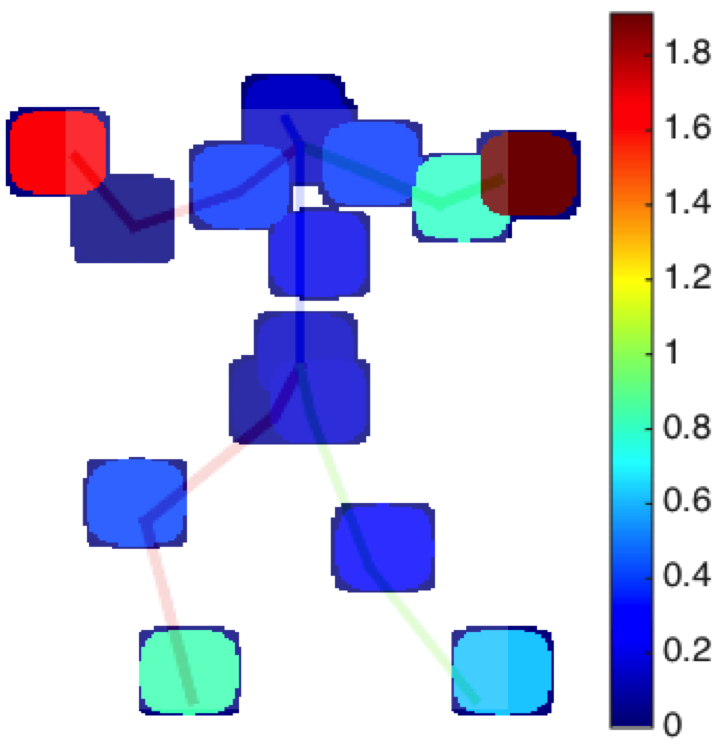}}	
	\subfigure[\scriptsize  3D pose:\newline 3D vs 3D+seg.+depth]{\includegraphics[height=2.8cm]{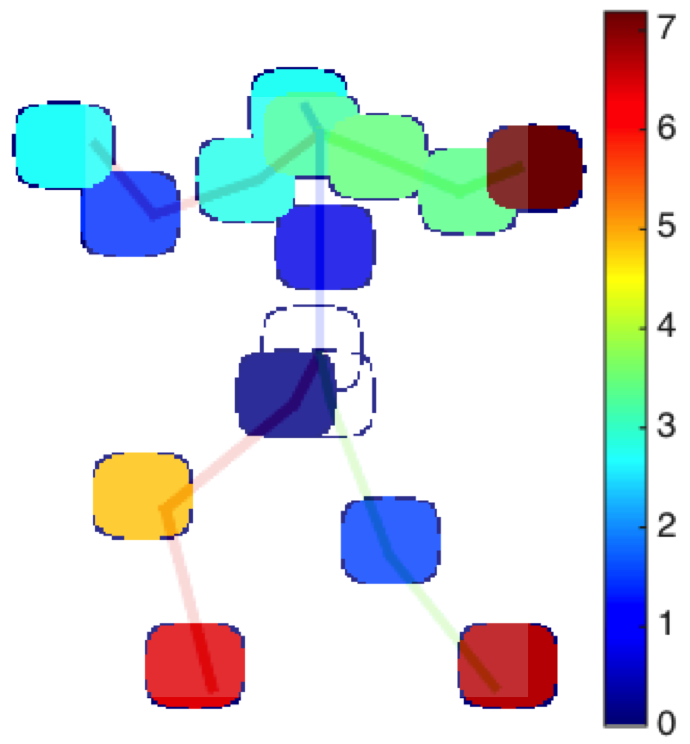}}	
	\vspace*{-2mm}
	\caption{Error visualization per each body part and task. The higher the value the higher the performance improvement for a particular metric of the best multi-task model compared to the baseline isolated task.}
	\label{fig:visual}
	\vspace*{-3mm}
\end{figure*}
	\subsubsection{3D pose estimation}
	
	This section analyzes the performance on 3D pose estimation of different multi-task models. Table~\ref{3dposetable} shows the prediction errors, in millimeters, for the different body joints and task combinations. The best overall results are obtained by considering the segmentation and depth estimation tasks along with 3D pose recovery, reducing the prediction error by $5\%$ (from $60.13$mm to $56.99$mm).
	
	It is interesting to see that, similarly to 2D pose recovery, where 3D pose did not help improve the predictions, now it is 3D pose that does not help. One can consider 2D pose recovery as a subtask of the 3D case, and thus the features used in 3D pose recovery already include those provided by the 2D case. In this case, the single task contributing the most to 3D pose recovery is segmentation, followed by depth estimation. This is likely due to the same reasons discussed in the previous section: providing an outline of the body parts, and providing a general outline of the body with depth information, helping to disambiguate between parts during pose recovery.
	
	Further combining both segmentation and depth estimation, as mentioned, obtains the best results, but not if we further consider 2D pose recovery. While in the previous section further adding 3D pose recovery to the 2D task did result in marginal benefits, in this case there is no further information provided: 2D landmark localization is a problem already tackled when performing the same task in the 3D space. This results in slightly worse results when considering all tasks: a larger feature representation is provided but without encoding extra information, facilitating over-fitting.

	If we inspect the results by body joint, we find the best combination of tasks for most joints includes segmentation and depth to the 3D pose. On the other hand, hips and thorax also benefit from including 2D body pose information. This is likely due to these parts forming the main portion of the body. A good 2D pose estimate may be more important for these parts, since the ambiguity in depth is smaller. For parts with more depth uncertainty, like the ankles, knees and wrists, considering 2D landmark estimation is highly detrimental to the 3D accuracy.
	
	\subsubsection{Analysis of success rate}
	We show success rate plots for different tasks in Fig. \ref{fig:success_rate}. For each modality we compare independent SH network with the best multi-task network performing that task. We also show the trend for one of the parts that multi-task approach better improves, specifically left wrist for 3D pose, Right ankle for 2D pose, left hand for depth map and left foot for part segmentation. As one can see in Fig. \ref{fig:success_rate}(c), full-body depth estimation benefits the most from multi-task learning, while 2D pose in Fig. \ref{fig:success_rate}(b) is the most accurate modality. In all cases, selected parts have higher than average gains for smaller error thresholds.
	

\subsection{Comparison to the state-of-the-art}
To the best of our knowledge, \cite{c23} is the only state-of-the-art multi-task work evaluating on SURREAL dataset. Similar to ours, they use SH modules to compute 2D/3D pose estimation and part segmentation. Differently from us, 2D pose and body part segmentation are independent streams feeding information to the 3D pose stream. Full body depth estimation is not considered. We compare the results in Table \ref{sotaTable}. Note that we exclude background to compute segmentation IoU as in \cite{c23}. Unlike \cite{c23} that trains 8 stacks for independent tasks and fine-tune 2 stacks in the multi-task model, we train our model from scratch using 2 stacks.

As one can see, our model is performing the best for 2D pose estimation in both independent and multi-task networks. Although our single-stream network performs better than \cite{c23} in segmentation, our multi-task approach obtains similar results. In the case of 3D pose estimation, \cite{c23} performs the best in both networks. Our multi-task network improves independent 3D pose by more than 3 mm while this improvement is 5.3 mm for \cite{c23}.

\begin{table}
    \centering
    \caption{State-of-the-art comparison on SURREAL}
    \vspace*{-3mm}
    \label{sotaTable}
    \begin{tabular}{|c|c|c|c|}
    \hline
    & Seg. & 2D pose & 3D pose \\ 
    & (IoU) & (PCKh) & (MJD mm) \\ \hline
    Varol \textit{et al.} \cite{c23} &  &  &  \\
    independent tasks & 59.2 & 82.7 & 46.1 \\ \hline
    Varol \textit{et al.} \cite{c23} &  &  &  \\
    multi-tasks & 69.2 & 90.8 & 40.8 \\ \hline
    Ours - independent tasks & 65.3 & 96.5 & 60.1 \\ \hline
    Ours - multi-tasks & 66.1 & 97.0 & 57.0 \\ \hline
    \end{tabular}
    \vspace*{-2mm}
\end{table}

\vspace*{-1mm}
\subsection{Discussion}

	This section summarizes some insights from the experiments performed for all tasks.
	
	We have seen that at the 2D level cues from depth estimation are highly useful for both body parts segmentation and human pose recovery, while 3D pose estimation contributes marginally to final performance. At the same time, body part segmentation and 2D pose estimation mutually benefit each other. Regarding body part segmentation, features from depth estimation improve the results the most, followed by 2D pose. Human pose recovery benefits from all other tasks, with the strongest cue being segmentation, followed by depth.
	
	In contrast, at the 3D level, depth estimation and human pose recovery benefit from segmentation, similarly to the two 2D tasks. In contrast, 2D pose cues are the least relevant, since we can interpret the task as a subtask of 3D pose recovery. Both tasks use the same model to get the lowest error, that is, depth + segmentation + 3D pose. We argue this is due to segmentation enriching the representation with semantic cues, and the extra depth information either providing a more restrictive deformation model (3D pose estimation) or a more dense depth representation (body depth estimation).
	
	Finally, a visual representation of the overall improvement of the best model per task and body part over the baseline is shown in Fig.~\ref{fig:visual}. The higher the value the better average improvement for each particular task metric (e.g. 1.2 for a 3D joint represents an average improvement of 1.2 MJD error reduction). We can see in IOU and Pixel accuracy that parts with more degrees of freedom, such as feet, hands and legs, are benefited the most from multi-tasking. In contrast, the trunk, head and upper arms, along with the background receive marginal improvements. For depth estimation, the improvements are more pronounced on the main body parts, such as the trunk and head, as well as the arms and hands. Then, for 2D and 3D pose, the former improved specially on the hands, while the latter improved on the upper body joints and ankles.

\section{Conclusions}
In this work we analyze the contribution of multi-tasking on four common body pose analysis problems: 2D/3D body pose recovery, full-body depth estimation and body parts segmentation. We have found that problems looking at complementary aspects of the problem benefit each other the most. Depth estimation and body part segmentation help each other, while 2D/3D body pose estimation benefit mainly from body part segmentation, followed by depth estimation. These tasks provide complementary features: depth information helps disambiguate body parts, while body part segmentation provides more robust features for locating joints during body pose estimation. Also, 3D pose estimation helps depth estimation, likely by reducing ambiguity: 3D pose estimation helps restrict the space of possible body poses. On the other hand, features from problems that are too closely related do not help significantly improve the predictions: 3D pose recovery already includes the 2D problem as a subtask, already encoding its features. For 2D pose recovery, features coming from the 3D case sacrifice precision in the camera plane, allotting more network capacity to estimate the landmarks depth.


\section{ACKNOWLEDGMENTS}
This work is partially supported by ICREA under the ICREA Academia programme, by the Spanish project TIN2016-74946-P (MINECO/FEDER, UE) and CERCA Programme / Generalitat de Catalunya. This work has been partially supported by the Spanish projects TIN2015-66951-C2-2-R (MINECO/FEDER, UE). We gratefully acknowledge the support of NVIDIA Corporation with the donation of the GPU used for this research.



\end{document}